# Exploiting compositionality to explore a large space of model structures

**Roger B. Grosse**
Comp. Sci. & AI Lab
MIT
Cambridge, MA 02139

**Ruslan Salakhutdinov**
Dept. of Statistics
University of Toronto
Toronto, Ontario, Canada

**William T. Freeman**
Comp. Sci. & AI Lab
MIT
Cambridge, MA 02139

**Joshua B. Tenenbaum**
Brain and Cognitive Sciences
MIT
Cambridge, MA 02193

## Abstract

The recent proliferation of richly structured probabilistic models raises the question of how to automatically determine an appropriate model for a dataset. We investigate this question for a space of matrix decomposition models which can express a variety of widely used models from unsupervised learning. To enable model selection, we organize these models into a context-free grammar which generates a wide variety of structures through the compositional application of a few simple rules. We use our grammar to generically and efficiently infer latent components and estimate predictive likelihood for nearly 2500 structures using a small toolbox of reusable algorithms. Using a greedy search over our grammar, we automatically choose the decomposition structure from raw data by evaluating only a small fraction of all models. The proposed method typically finds the correct structure for synthetic data and backs off gracefully to simpler models under heavy noise. It learns sensible structures for datasets as diverse as image patches, motion capture, 20 Questions, and U.S. Senate votes, all using exactly the same code.

## 1 Introduction

There has been much interest recently in learning hierarchical models, which extend simpler models by introducing additional dependencies between the parameters. While there have been many advances in modeling particular kinds of structure, as the desired structure becomes higher level and more abstract, the correct model becomes less obvious a priori. We aim to determine an appropriate model structure automatically from the data, in order to make hierarchical modeling usable by non-experts and to explore a wider variety of structures than would be possible by manual engineering.

There has been much work on structure learning in particular model classes, such as undirected (Lee et al., 2006) and directed (Teyssier and Koller, 2005) graphical models. Most such work focuses on determining the particular factorization and/or conditional independence structure within a fixed model class. Our concern, however, is with identifying the overall form of the model. For instance, suppose we are interested in modeling the voting patterns of U.S. Senators. We can imagine several plausible modeling assumptions for this domain: e.g., that political views can be summarized by a small number of dimensions, that Senators cluster into voting blocks, or that votes can be described in terms of binary attributes. Choosing the correct assumptions is crucial for uncovering meaningful structure. Right now, the choice of modeling assumptions is heavily dependent on the intuition of the human researcher; we are interested in determining appropriate modeling assumptions automatically.

Many common modeling assumptions, or combinations thereof, can be expressed by a class of probabilistic models called matrix decompositions. In a matrix decomposition model, component matrices are first sampled independently from a small set of priors, and then combined using simple algebraic operations. This expressive model class can represent a variety of widely used models, including clustering, co-clustering (Kemp et al., 2006), binary latent factors (Griffiths and Ghahramani, 2005), and sparse coding (Olshausen and Field, 1996). Nevertheless, the space of models is *compositional*: each model is described recursively in terms of simpler matrix decomposition models and the operations used to combine them. We propose to exploit this compositional structure to efficiently and generically evaluate and perform inference in matrix decomposition models, and to automatically search through the space of structures to find one appropriate for a dataset.

A common heuristic researchers use for designing hierarchical models is to fit an existing model, look for additional dependencies in the learned representation, and extend the model to capture those dependencies. We formalize this process in terms of a context-free grammar. In particular, we present a notation for describing matrix decomposition models as algebraic expressions, and organize the space of models into a context-free grammar which generates such expressions. The starting symbol corresponds to a structureless model where the entries of the input matrix are modeled as *i.i.d.* Gaussians. Each production rule corresponds to a simple unsupervised learning model, such as clustering or dimensionality reduction. These production rules lie at the heart of our approach: we fit and evaluate a wide variety of models using a small toolbox of algorithms corresponding to the production rules, and the production rules guide our search through the space of structures.

The main contributions of this paper are threefold. First, we present a unifying framework for matrix decompositions based on a context-free grammar which generates a wide variety of structures through the compositional application of a few simple production rules. Second, we exploit our grammar to infer the latent components and estimate predictive likelihood in all of these structures generically and efficiently using a small toolbox of reusable algorithms corresponding to different component matrix priors and production rules. Finally, by performing greedy search over our grammar using predictive likelihood as the criterion, we can (in practice) typically choose the correct structure from the data while evaluating only a small fraction of all possible structures.

Section 3 defines our matrix decomposition formalism, and Sections 4 and 5 present our generic algorithms for inferring component matrices and evaluating individual structures, respectively. Section 6 describes how we search our space of structures. Finally, in Section 7, we evaluate the structure learning procedure on synthetic data and on real-world datasets as diverse as image patches, motion capture, 20 Questions, and Senate voting records, all using *exactly the same code*. Our procedure learns correct and/or plausible model structures for a wide variety of synthetic and real datasets, and gracefully falls back to simpler structures in high-noise conditions.

## 2 Related work

There is a long history of attempts to infer model structures automatically. The field of algorithmic information theory (Li and Vitanyi, 1997) studies how to represent data in terms of a short program/input pair which could have generated it. One prominent example, Solomonoff induction, can learn any computable generative model, but is itself uncomputable. Minimum message length (Wallace, 2005), minimum description length (Barron et al., 1998), and Bayesian model comparison (MacKay, 1992) are frameworks which can, in principle, be used to compare very different generative models. In practice, they have primarily been used for controlling complexity within a given model class. By contrast, our aim is to choose from a very large space of model classes by exploiting shared structure between the models.

Other work has focused on searching within more restricted spaces of models, such as undirected (Lee et al., 2006) and directed (Teyssier and Koller, 2005) graphical models, and graph embeddings (Kemp and Tenenbaum, 2008). Kemp and Tenenbaum (2008) model human "domain structure" learning as selecting between a fixed set of graph structures. Similarly to this paper, their structures are generated from a few simple rules; however, whereas their set of structures is small enough to exhaustively evaluate each one, we search over a much larger set of structures in a way that explicitly exploits the recursive nature of the space. Furthermore, our space of matrix decomposition structures is especially broad, including many bread-and-butter models from unsupervised learning, as well as the building blocks of many hierarchical Bayesian models.

We note that several other researchers have proposed unifying frameworks for unsupervised learning which overlap substantially with our own. Roweis and Ghahramani (1999)'s "generative model for generative models" presents a lattice showing relationships between different models. Srebro (2004) and Singh and Gordon (2008) each interpreted a variety of unsupervised learning algorithms as factorizing an input matrix into a product of two factors. Exponential family PCA (Collins et al., 2002; Mohamed et al., 2008) generalizes low-rank factorizations to other observation models in the exponential family. Our work differs from these in that our matrix decomposition formalism is specifically designed to support efficient generic inference and structure learning. We defer discussion of particular matrix decomposition models to Section 3.1, after we have introduced our formalism.

Our work has several parallels in the field of equation discovery. Langley et al. (1984) built a knowledge discovery system called BACON which reproduced classical scientific discoveries. BACON followed a greedy search procedure similar to our own: it repeatedly fit statistical models and looked for additional structure in the learned parameters. Our work is similar in spirit, but uses matrices rather than scalars as the building blocks, allowing us to capture rich structure in high-dimensional spaces. Todorovski and Dzeroski (1997) used a context-free grammar to define spaces of candidate equations. Our approach differs in that we explicitly use the grammar to structure posterior inference and search over model structures.

## 3 A grammar for matrix decompositions

We first present a notation for describing matrix decomposition models as algebraic expressions, such as $MG + G$. Each letter corresponds to a particular distribution over matrices. When the dimensions of all the component matrices are specified, the algebraic expression defines a generative model: first sample all of the component matrices independently from their corresponding priors, and then evaluate the expression. The component priors are as follows:

1. **Gaussian (G).** Entries are independent Gaussians:

$$u_{ij} \sim \text{Gaussian}(0, \lambda_i^{-1}\lambda_j^{-1}).$$

This is our most generic component prior, and gives a way of deferring or ignoring structure.[1]

2. **Multinomial (M).** Rows are independent multinomials, with one 1 and the rest 0's:

$$\pi \sim \text{Dirichlet}(\alpha) \qquad u_i \sim \text{Multinomial}(\pi).$$

This is useful for clustering models, where $u_i$ determines the cluster assignment for the $i^{th}$ row.

3. **Bernoulli (B).** Entries are independent Bernoullis:

$$\pi_j \sim \text{Beta}(a, b) \qquad u_{ij} \sim \text{Bernoulli}(\pi_j).$$

This is useful for binary latent feature models.

4. **Integration matrix (C).** Entries below the diagonal are deterministically 1:

$$u_{ij} = \mathbf{1}_{i \geq j}.$$

This is useful for modeling temporal structure, as multiplying by this matrix has the effect of cumulatively summing the rows. (Mnemonic: C for "cumulative.")

We allow expressions consisting of addition, matrix multiplication, matrix transpose, elementwise multiplication ($\circ$), and elementwise exponentiation ($\exp$). Some of the dimensions of the component matrices are determined by the size of the input matrix; the rest (the latent dimensions) are determined automatically using the techniques of Section 4.

We observe that this notation for matrix decompositions is recursive: each sub-expression (such as $GM^T + G$ in the above example) is itself a matrix decomposition model. Furthermore, the semantics is compositional: the value of each expression depends only on the values of its sub-expressions and the operations used to combine them. These observations motivate our decision to define a space of models using a context-free grammar, a formalism which is widely used for representing recursive and compositional structures such as languages.

---
[1]The precision parameters $\lambda_i$ and $\lambda_j$ are drawn from the distribution $\text{Gamma}(a, b)$. If $i$ indexes a data dimension (i.e. rows correspond to rows of the input matrix), the $\lambda_i$s are tied. This allows the variance parameters to generalize to additional rows. If $i$ indexes a latent dimension, the $\lambda_i$s are all independent draws. This allows the variances of latent dimensions to be estimated. The same holds for the $\lambda_j$s.

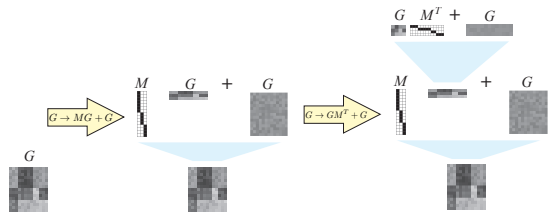

Figure 1: A synthetic example showing how an input matrix with block structure can be co-clustered by fitting the matrix decomposition structure $M(GM^T + G) + G$. Rows and columns are sorted for visualization purposes.

The starting symbol in our grammar is $G$, a structureless model where the entries are assumed to be independent Gaussians. Other models (expressions) are generated by repeatedly applying one of the following production rules:

| | | |
|---:|:---|---:|
| low-rank approximation | $G \to GG + G$ | (1) |
| clustering | $G \to MG + G \mid GM^T + G$ | (2) |
| | $M \to MG + G$ | (3) |
| linear dynamics | $G \to CG + G \mid GC^T + G$ | (4) |
| sparsity | $G \to \exp(G) \circ G$ | (5) |
| binary factors | $G \to BG + G \mid GB^T + G$ | (6) |
| | $B \to BG + G$ | (7) |
| | $M \to B$ | (8) |

For instance, any occurrence of $G$ in a model may be replaced by $GG + G$ or $MG + G$. Repeated application of these production rules allows us to build hierarchical models by capturing additional dependencies between variables which were previously modeled as independent.

### 3.1 Examples

We now turn to several examples in which our simple components and production rules give rise to a rich variety of models from unsupervised learning. While the model space is symmetric with respect to rows and columns, for purposes of exposition, we will adopt the convention that the rows of the input matrix correspond to data points and columns corresponds to observed attributes.

We always begin with the model $G$, which assumes the entries of the matrix are *i.i.d.* Gaussian. Applying productions in our grammar allows us to capture additional structure. For instance, starting with Rule 2(a) gives the model $MG + G$, which clusters the rows (data points). In more detail, the $M$ represents the cluster assignments, the first $G$ represents the cluster centers, and the second $G$ represents within-cluster variation. These three matrices are sampled independently, the assignment matrix is multiplied by the center matrix, and the within-cluster variation is added to the result. By applying Rule 2(b), the clustering model can be extended to co-clustering (Kemp et al., 2006), where the columns (attributes) form clusters as well. In our framework, this can be represented as $M(GM^T + G) + G$. We need not stop here: for instance, there may be coherent co-variation even within individual clusters. One can capture this variation by applying Rule 3 to get the Bayesian Clustered Tensor Factorization (BCTF) (Sutskever et al., 2009) model $(MG+G)(GM^T+G)+G$. This process is shown in cartoon form in Figure 1.

For an example from vision, consider a matrix $X$, where each row is a small (e.g. $12 \times 12$) patch sampled from an image and vectorized. Image patches can be viewed as lying near a low-dimensional subspace spanned by the lowest frequency Fourier coefficients (Bossomaier and Snyder, 1986). This can be captured by the low-rank model $GG+G$. In a landmark paper, Olshausen and Field (1996) found that image patches are better modeled as a linear combination of a small number of components drawn from a larger dictionary. In other words, $X$ is approximated as the product $WA$, where each row of $A$ is a basis function, and $W$ is a sparse matrix giving the linear reconstruction coefficients for each patch. By fitting this "sparse coding" model, they obtained a dictionary of oriented edges similar to the receptive fields of neurons in the primary visual cortex. If we apply Rule (5), we obtain a Bayesian version of sparse coding, $(\exp(G) \circ G)G+G$, similar to the model proposed by Berkes et al. (2008). Intuitively, the latent Gaussian coefficients are multiplied elementwise by "scale" variables to give a heavy-tailed distribution. Many researchers have designed models to capture the dependencies between these scale variables, and such "Gaussian scale mixture" models represent the state-of-the art for low-level vision tasks such as denoising (Portilla et al., 2003) and texture synthesis (Portilla and Simoncelli, 2000). One such GSM model is that of Karklin and Lewicki (2008), who fit a low-rank model to the scale variables. By applying Rule (1) to the sparse coding structure, we can represent their model in our framework as $(\exp(GG+G)\circ G)G+G$. This model has been successful at capturing higher-level textural properties of a scene and has properties similar to complex cells in the primary visual cortex.

Figure 2 gives several additional examples of matrix decomposition models and highlights the relationships between them. We emphasize that our goal is not to reproduce existing models exactly, but to develop a formalism powerful enough to express a wide variety of statistical assumptions about the latent factors underlying the data.

We note that many of the above models are not typically viewed as matrix decomposition structures. Describing them as such results in a compact notation for defining them and makes clearer the relationships between the different models. The above examples have in common that complex models can be derived by incrementally adding structure to a sequence of simpler models (in a way that parallels the path researchers took to discover them). This observation motivates our proposed procedures for inference and structure learning.

## 4 Posterior inference of component matrices

Searching over matrix decomposition structures requires a generic and unified approach for posterior sampling of the latent matrices. Unfortunately, for most of the structures we consider, this posterior is complicated and multimodal, and escaping from local modes requires carefully chosen special-purpose sampling operators. Engineering such operators for thousands of different models would be undesirable.

Fortunately, the compositional nature of our model space allows us to focus the engineering effort on the relatively small number of production rules. In particular, observe that in a realization of the generative process, the value of an expression depends only on the values of its sub-expressions. This suggests the following initialization procedure: when applying a production rule $P$ to a matrix $S$, sample from the posterior for $P$'s generative model conditioned on it evaluating (exactly) to $S$. Many of our production rules correspond to simple machine learning models for which researchers have already expended much time developing efficient inference algorithms:

1. **Low rank.** To apply the rule $G \to GG+G$, we fit the probabilistic matrix factorization (Salakhutdinov and Mnih, 2008) model using block Gibbs sampling over the two factors. While PMF assumes a fixed latent dimension, we choose the dimension automatically by placing a Poisson prior on the dimension and moving between states of differing dimension using reversible jump MCMC (Green, 1995).

2. **Clustering.** To apply the clustering rule to rows: $G \to MG+G$, or to columns: $G \to GM^T+G$, we perform collapsed Gibbs sampling over the cluster assignments in a Dirichlet process mixture model.

3. **Binary factors.** To apply the rule $G \to BG+G$ or $G \to GB^T+G$, we perform accelerated collapsed Gibbs sampling (Doshi-Velez and Ghahramani, 2009) over the binary variables in a linear-Gaussian Indian Buffet Process (Griffiths and Ghahramani, 2005) model, using split-merge proposals (Meeds et al., 2006) to escape local modes.

4. **Markov chains.** The rule $G \to CG+G$ is equivalent to estimating the state of a random walk given noisy observations, which is done using Rauch-Tung-Striebel (RTS) smoothing.

The remaining production rules begin with a random decomposition of $S$. While some of these algorithms involve fitting Bayesian nonparametric models, once the dimensionality is chosen, the model is converted to a finite model of fixed dimensionality (as defined in section 3). The smart initialization step is followed by generic Gibbs sampling over the entire model. We note that our initialization procedure generalizes "tricks of the trade" whereby complex models are initialized from simpler ones (Kemp et al., 2006; Miller et al., 2009).

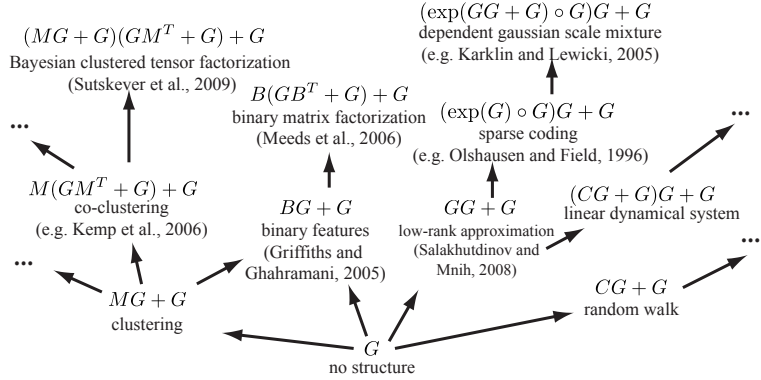

Figure 2: Examples of existing machine learning models which fall under our framework. Arrows represent models reachable using a single production rule. Only a small fraction of the 2496 models reachable within 3 steps are shown, and not all possible arrows are shown.

In addition to simplifying the engineering, this procedure allows us to reuse computations between different structures. Most of the computation time is in the initialization steps. Each of these steps only needs to be run once on the full matrix, specifically when the first production rule is applied. Subsequent initialization steps are performed on the component matrices, which are considerably smaller. This allows a large number of high level structures to be fit for a fraction of the cost of fitting them from scratch.

## 5 Scoring candidate structures

Performing model selection requires a criterion for scoring individual structures which is informative yet tractable. To motivate our method, we first discuss two popular choices: marginal likelihood of the input matrix and entrywise mean squared error (MSE). Marginal likelihood, the probability of the data with all the latent variables integrated out, is widely used in Bayesian model selection. Unfortunately, this requires integrating out all of the latent component matrices, whose posterior distribution is highly complex and multimodal. While elegant solutions exist for particular models, estimating the data marginal likelihood generically is still extremely difficult. At the other extreme, one can hold out a subset of the entries of the matrix and compute the mean squared error for predicting these entries. MSE is easier to implement, but we found that it was unable to distinguish many of the the more complex structures in our grammar.

As a compromise between these two approaches, we chose to evaluate predictive likelihood of held-out rows and columns. That is, for each row (or column) $x$ of the matrix, we evaluate $p(x|X_O)$, where $X_O$ denotes an "observed" sub-matrix. Like marginal likelihood, this tests the model's ability to predict entire rows or columns. However, it can be efficiently approximated in our class of models using a small but carefully chosen toolbox corresponding to the component matrix priors in our grammar. We discuss the case of held-out rows; columns are handled analogously.

First, by expanding out the products in the expression, we can write the decomposition uniquely in the form

$$X = U_1 V_1 + \cdots + U_n V_n + E, \qquad (1)$$

where $E$ is an *i.i.d.* Gaussian "noise" matrix and the $U_i$'s are any of the following: (1) a component matrix $G$, $M$, or $B$, (2) some number of $C$s followed by $G$, (3) a Gaussian scale mixture. The held-out row $x$ can therefore be represented as:

$$x = V_1^T u_1 + \cdots + V_n^T u_n + e. \qquad (2)$$

The predictive likelihood is given by:

$$p(x|X_O) = \int p(U_O, V|X_O) p(u|U_O) p(x|u, V) \, dU_O \, du \, dV \qquad (3)$$

where $U_O$ is shorthand for $(U_{O1}, \ldots, U_{On})$ and $u$ is shorthand for $(u_1, \ldots, u_n)$.

In order to evaluate this integral, we generate samples from the posterior $p(U_O, V|X)$ using the techniques described in Section 4, and compute the sample average of

$$p_{pred}(x) \triangleq \int p(u|U_O) p(x|u, V) \, du \qquad (4)$$

If the term $U_i$ is a Markov chain, the predictive distribution $p(u_i|U_O)$ can be computed using Rauch-Tung-Striebel smoothing; in the other cases, $u$ and $U_O$ are related only through the hyperparameters of the component prior. Either way, each term $p(u_i|U_O)$ can be summarized as a Gaussian, multinomial, Bernoulli, or Gaussian scale mixture distribution.

It remains to marginalize out the latent representation $u$ of the held-out row. While this can be done exactly in some simple models, it is intractable in general (for instance, if $u$ is Bernoulli or a Gaussian scale mixture). It is important that the approximation to the integral be a lower bound, because otherwise an overly optimistic model could be chosen even when it is completely inappropriate.

Our approach is a hybrid of variational and sampling techniques. We first lower bound the integral (4) in an approximate model $\tilde{p}$ where the Gaussian scale mixture components are approximated as Gaussians. This is done using using the variational Bayes bound

$$\log \tilde{p}_{pred}(x) \geq E_q[\log \tilde{p}_{pred}(x, u)] + \mathcal{H}(q).$$

The approximating distribution $q(u)$ is such that all of the discrete components are independent, while the Gaussian components are marginalized out. The ratio $p_{pred}(x)/\tilde{p}_{pred}(x)$ is then estimated using annealed importance sampling (AIS) (Neal, 2001). Because AIS is an unbiased estimator which always takes positive values, by Markov's inequality we can regard it as a stochastic lower bound. Therefore, this small toolbox of techniques allows us to (stochastically) lower bound the predictive likelihood across a wide variety of matrix decomposition models.

## 6 Search over structures

We aim to find a matrix decomposition structure which is a good match to a dataset, as measured by the predictive likelihood criterion of Section 5. Since the space of models is large and inference in many of the models is expensive, we wish to avoid exhaustively evaluating every model. Instead, we adopt a greedy search procedure inspired by the process of scientific discovery. In particular, consider a common heuristic researchers use to build probabilistic models: we begin with a model which has already been applied to a problem, look for additional dependencies not captured by the model, and refine the model to account for those dependencies.

In our approach, refining a model corresponds to applying one of the productions. This suggests the following greedy search procedure, which iteratively "expands" the best-scoring unexpanded models by applying all possible production rules and scoring the resulting models. In particular we first expand the structureless model $G$. Then, in each step, we expand the $K$ best-performing models from the previous step by applying all possible productions. We then score all the resulting models. The procedure stops when no model achieves sufficient improvement over the best model from the previous step. We refer to the models reached in $i$ productions as the Level $i$ models; for instance, $GG+G$ is a Level 1 model and $(MG+G)G+G$ is a Level 2 model.

The effectiveness of this search procedure depends whether the score of a simple structure is a strong indicator of the best score which can be obtained from the structures derived from it. In our experiments, the scores of the simpler structures turned out to be a powerful heuristic: while our experiments used $K = 3$, in most all cases, the correct (or best-scoring) structure would have been found with a purely greedy search ($K = 1$). This results in enormous savings because of the compositional nature of our search space: while the number of possible structures (up to a given level) grows quickly in the number of production rules, the number of structures evaluated by this search procedure is merely linear.

The search procedure returns a high-scoring structure for each level in our grammar. There remains a question of when to stop. Choosing between structures of differing complexity imposes a tradeoff between goodness of fit and other factors such as interpretability and tractability of inference, and inevitably the choice is somewhat subjective. In practice, a user may wish to run our procedure up to a fixed level and analyze the sequence of models chosen, as well as the predictive likelihood improvement at each level. However, for the purposes of evaluating our system, we need it to return a single answer. In all of our experiments, we adopt the following arbitrary but consistent criterion: prefer the higher level structure if its predictive log-likelihood score improves on the previous level by at least one nat per row and column.[2]

## 7 Experiments

### 7.1 Synthetic data

We first validated our structure learning procedure on synthetic data where the correct model was known. We generated matrices of size $200 \times 200$ from all of the models in Figure 2, with 10 latent dimensions. The noise variance $\sigma^2$ was varied from 0.1 to 10, while the signal variance was fixed at 1.[3] The structures selected by our procedure are shown in Table 1.

---

[2]More precisely, if $\frac{S_i - S_{i-1}}{N+D} > 1$, where $S_i$ is the total predictive log-likelihood for the level $i$ model summed over all rows and columns, and $N$ and $D$ are the numbers of rows and columns, respectively. We chose to normalize by $N+D$ because the predictive likelihood improvements between more abstract models tend to grow with the number of rows and columns in the input matrix, rather than the number of entries.

[3]Our grammar generates expressions of the form $\cdots + G$. We consider this final $G$ term to be the "noise" and the rest to be the "signal," even though the models and algorithms do not distinguish the two.

|  | — Increasing noise ⟶ | | | |
| ---: | --- | --- | --- | --- |
|  | $\sigma^2 = 0.1$ | $\sigma^2 = 1$ | $\sigma^2 = 3$ | $\sigma^2 = 10$ |
| low-rank | $GG + G$ | $GG + G$ | $GG + G$ | ①$G$ |
| clustering | $MG + G$ | $MG + G$ | $MG + G$ | $MG + G$ |
| binary latent features | ①$(BG + G)G + G$ | $BG + G$ | $BG + G$ | $BG + G$ |
| co-clustering | $M(GM^T + G) + G$ | $M(GM^T + G) + G$ | $M(GM^T + G) + G$ | ①$GM^T + G$ |
| binary matrix factorization | ①$(BG + G)(GB^T + G) + G$ | $(BG + G)B^T + G$ | ②$GG + G$ | ②$GG + G$ |
| BCTF | $(MG + G)(GM^T + G) + G$ | $(MG + G)(GM^T + G) + G$ | ②$GM^T + G$ | ●$G$ |
| sparse coding | $(\exp(G) \circ G)G + G$ | $(\exp(G) \circ G)G + G$ | $(\exp(G) \circ G)G + G$ | ②$G$ |
| dependent GSM | ①$(\exp(G) \circ G)G + G$ | ①$(\exp(G) \circ G)G + G$ | ①$(\exp(G) \circ G)G + G$ | ●$BG + G$ |
| random walk | $CG + G$ | $CG + G$ | $CG + G$ | ①$G$ |
| linear dynamical system | $(CG + G)G + G$ | $(CG + G)G + G$ | $(CG + G)G + G$ | ②$BG + G$ |

Table 1: The structures learned from $200 \times 200$ matrices generated from various distributions, with signal variance 1 and noise variance $\sigma^2$. Incorrect structures are marked with a 1, 2, or 3, depending how many decisions would need to be changed to find the correct structure. We observe that our approach typically finds the correct answer in low noise settings and backs off to simpler models in high noise settings.

We observe that seven of the ten structures were identified perfectly in both trials where the noise variance was no larger than the data variance ($\sigma^2 \leq 1$). When $\sigma^2 = 0.1$, the system incorrectly chose $(BG+G)G+G$ for the binary latent feature data, rather than $BG + G$. Similarly, it chose $(BG+G)(GB^T+G)+G$ rather than $(BG+G)B^T+G$ for binary matrix factorization. In both cases, the sampler learned an incorrect set of binary features, and the additional flexibility of the more complex model compensated for this. This phenomenon, where more structured models compensate for algorithmic failures in simpler models, has also been noted in the context of deep learning (Salakhutdinov and Murray, 2008).

Our system also did not succeed in learning the dependent Gaussian scale mixture structure $(\exp(GG+G)\circ G)G+G$ from synthetic data, instead generally falling back to the simpler sparse coding model $(\exp(G) \circ G)G + G$. For $\sigma^2 = 0.1$ the correct structure was in fact the highest scoring structure, but did not cross our threshold of 1 nat improvement over the previous level. We note that in every case, there were nearly 2500 incorrect structures to choose from, so it is notable that the correct model structure can be recovered most of the time.

In general, when the noise variance was much larger than the signal variance, the system gracefully fell back to simpler models, such as $GM^T +G$ instead of the BCTF model $(MG+G)(GM^T +G)+G$ (see Section 3.1). At the extreme, in the maximum noise condition, it chose the structureless model $G$ much of the time. Overall, our procedure reliably learned most of the model structures in low-noise settings (impressive considering the extremely large space of possible wrong answers) and gracefully fell back to simpler models when necessary.

### 7.2 Real-world data

Next, we evaluated our system on several real-world datasets. We first consider two domains, motion capture and image statistics, where the core statistical assumptions are widely agreed upon, and verify that our learned structures are consistent with these assumptions. We then turn to domains where the correct structure is more ambiguous and analyze the representations our system learns.

In general, we do not expect every real-world dataset to have a unique best structure. In cases where the predictive likelihood score differences between multiple top-scoring models were not statistically significant, we report the set of top-scoring models and analyze what they have in common.

**Motion capture.** We first consider a human motion capture dataset (Hsu et al., 2005; Taylor et al., 2007) consisting of a person walking in a variety of styles. Each row of the matrix gives the person's orientation and displacement in one frame, as well as various joint angles. We used 200 frames (6.7 seconds), and 45 state variables. In the first step, the system chose the Markov chain model $CG + G$, which assumes that the components of the state evolve continuously but independently over time. Since a person's different joint angles are clearly correlated, the system next captured these correlations with the model $C(GG+G)+G$. This is slightly different from the popular linear dynamical system model $(CG + G)G + G$, but it is more physically correct in the sense that the LDS assumes the deviations of the observations from the low-dimensional subspace must be independent in different time steps, while our learned structure captures the temporal continuity in the deviations.

**Natural image patches.** We tested the system on the Sparsenet dataset of Olshausen and Field (1996), which consists of 10 images of natural scenes which were blurred and whitened. The rows of the input matrix corresponded to 1,000 patches of size $12 \times 12$. In the first stage, the model learned the low-rank representation $GG + G$, and in the second stage, it sparsified the linear reconstruction coefficients to give the sparse coding model $(\exp(G)\circ G)G+G$. In the third round, it modeled the dependencies between the scale variables by recursively giving them a low-rank representation, giving a dependent Gaussian scale mixture (GSM) model $(\exp(GG + G) \circ G)G + G$ reminiscent of Karklin and Lewicki (2008). A closely related model, $(\exp(GB^T + G) \circ G)G + G$, also achieved a score not significantly lower. Both of these structures resulted in a rank-one factorization of the scale matrix, similar to the radial Gaussianization model of Lyu and Simoncelli (2009) for neighboring wavelet coefficients.

|  | Level 1 | Level 2 | Level 3 |
|---|---|---|---|
| Motion capture | $CG + G$ | $C(GG + G) + G$ | — |
| Image patches | $GG + G$ | $(\exp(G) \circ G)G + G$ | $(\exp(GG + G) \circ G)G + G$ |
| 20 Questions | $MG + G$ | $M(GG + G) + G$ | — |
| Senate votes | $GM^T + G$ | $(MG + G)M^T + G$ | — |

Table 2: The best performing models at each level of our grammar for real-world datasets. These correspond to plausible structures for the datasets, as discussed in the text.

Dependent GSM models (see Section 3.1) are the state-of-the-art for a variety of image processing tasks, so it is interesting that this structure can be learned merely from the raw data. We note that a single pass through the grammar reproduces an analogous sequence of models to those discovered by the image statistics research community as discussed in Section 3.1.

**20 Questions.** We now consider a dataset collected by Pomerleau et al. (2009) of Mechanical Turk users' responses to 218 questions from the 20 Questions game about 1000 concrete nouns (e.g. animals, foods, tools). The system began by clustering the entities using the flat clustering model $MG + G$. In the second stage, it found low-rank structure in the matrix of cluster centers, resulting in the model $M(GG + G) + G$. No third-level structure achieved more than 1 nat improvement beyond this. The low-rank representation had 8 dimensions, where the largest variance dimension corresponded to living vs. nonliving and the second largest corresponded to large vs. small. The 39 clusters, the 20 largest of which are shown in Figure 3, correspond to semantically meaningful categories.

We note that two other models expressing similar assumptions, $M(GB^T + G) + G$ and $(MG + G)G + G$, achieved scores only slightly lower. What these models have in common is a clustering of entities (but not questions) coupled with low-rank structure between entities and questions. The learned clusters and dimensions are qualitatively similar in each case.

**Senate voting records.** Finally, we consider a dataset of roll call votes from the 111th United States Senate (2009-2010). Rows correspond to Senators, and the columns correspond to all 696 votes, most of which were on procedural motions and amendments to bills. Yea votes were mapped to 1, Nay and Present were mapped to -1, and absences were treated as unobserved. In the first two stages, our procedure clustered the votes and Senators, giving the clustering model $GM^T + G$ and the co-clustering model $(MG + G)M^T + G$, respectively. Senators clustered along party lines, as did most of the votes, according to the party of the proposer. The learned representations are all visualized in Figure 4.

In the third stage, one of the best performing models was Bayesian clustered tensor factorization (BCTF) (see section 3.1), where Senators and votes are each clustered inside a low-rank representation.[4] This low-rank representation was rank 5, with one dominant dimension corresponding to the liberal-conservative axis. The BCTF model makes it clearer that the clusters of Senators and votes are not independent, but can be seen as occupying different points in a low-dimensional representation. This model improved on the previous level by less than our 1 nat cutoff.[5] The models in this sequence increasingly highlight the polarization of the Senate.

## 8 Discussion

We have presented an effective and practical method for automatically determining the model structure in a particular space of models, matrix decompositions, by exploiting compositionality. However, we believe our approach can be extended beyond the particular space of models presented here. Most straightforwardly, additional components can be added to capture other motifs of probabilistic modeling, such as tree embeddings and low-dimensional embeddings. More generally, it should be fruitful to investigate other model classes with compositional structure, such as tensor decompositions.

In either case, exploiting the structure of the model space becomes increasingly essential. For instance, the number of models reachable in 3 steps is cubic in the number of production rules, whereas the complexity of the greedy search is linear. For tensors, the situation is even more overwhelming: even if we restrict our attention to analogues of $GG+G$, a wide variety of provably distinct generalizations have been identified, including the widely used Tucker3 and PARAFAC decompositions (Kolda and Bader, 2007).

---

[4] The other models whose scores were not significantly different were: $(MG+G)M^T+BG+G$, $(MG+G)M^T+GM^T+G$, $G(GM^T+G)+GM^T+G$, and $(BG+G)(GM^T+G)+G$. All of these models include the clustering structure but account for additional variability within clusters.

[5] BCTF results in a more compact representation than the co-clustering model, but our predictive likelihood criterion doesn't reward this except insofar as overfitting hurts a model's ability to generalize to new rows and columns. We speculate that a fully Bayesian approach using marginal likelihood may lead to more compact structures.

1. **Miscellaneous.** key, chain, powder, aspirin, umbrella, quarter, cord, sunglasses, toothbrush, brush
2. **Clothing.** coat, dress, pants, shirt, skirt, backpack, tshirt, quilt, carpet, pillow, clothing, slipper, uniform
3. **Artificial foods.** pizza, soup, meat, breakfast, stew, lunch, gum, bread, fries, coffee, meatballs, yoke
4. **Machines.** bell, telephone, watch, typewriter, lock, channel, tuba, phone, fan, ipod, flute, aquarium
5. **Natural foods.** carrot, celery, corn, lettuce, artichoke, pickle, walnut, mushroom, beet, acorn
6. **Buildings.** apartment, barn, church, house, chapel, store, library, camp, school, skyscraper
7. **Printed things.** card, notebook, ticket, note, napkin, money, journal, menu, letter, mail, bible
8. **Body parts.** arm, eye, foot, hand, leg, chin, shoulder, lip, teeth, toe, eyebrow, feet, hair, thigh
9. **Containers.** bottle, cup, glass, spoon, pipe, gallon, pan, straw, bin, clipboard, carton, fork
10. **Outdoor places.** trail, island, earth, yard, town, harbour, river, planet, pond, lawn, ocean
11. **Tools.** knife, chisel, hammer, pliers, saw, screwdriver, screw, dagger, spear, hoe, needle
12. **Stuff.** speck, gravel, soil, tear, bubble, slush, rust, fat, garbage, crumb, eyelash
13. **Furniture.** bed, chair, desk, dresser, table, sofa, seat, ladder, mattress, handrail, bench, locker
14. **Liquids.** wax, honey, pint, disinfectant, gas, drink, milk, water, cola, paste, lemonade, lotion
15. **Structural features.** bumper, cast, fence, billboard, guardrail, axle, deck, dumpster, windshield
16. **Non-solid things.** surf, fire, lightning, sky, steam, cloud, dance, wind, breeze, tornado, sunshine
17. **Transportation.** airplane, car, train, truck, jet, sedan, submarine, jeep, boat, tractor, rocket
18. **Herbivores.** cow, horse, lamb, camel, pig, hog, calf, elephant, cattle, giraffe, yak, goat
19. **Internal organs.** rib, lung, vein, stomach, heart, brain, smile, blood, lap, nerve, lips, wink
20. **Carnivores.** bear, walrus, shark, crocodile, dolphin, hippo, gorilla, hyena, rhinocerous

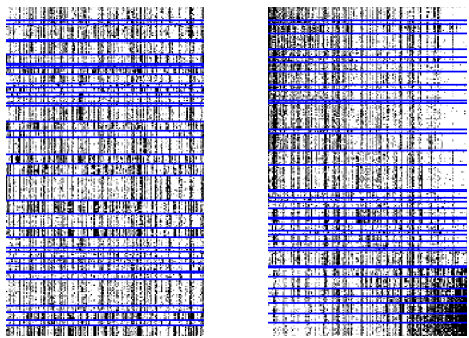

Figure 3: **(left)** The 20 largest clusters discovered by our Level 2 model $M(GG + G) + G$ for the 20 Questions dataset. Each line gives **our interpretation**, followed by random items from the cluster. **(right)** Visualizations of the Level 1 representation $MG + G$ and the Level 2 representation $M(GG + G) + G$. Rows = entities, columns = questions. 250 rows and 150 columns were selected at random from the original matrix. Rows and columns are sorted first by cluster, then by the highest variance dimension of the low-rank representation (if applicable). Clusters were sorted by the same dimension as well. Blue = cluster boundaries.

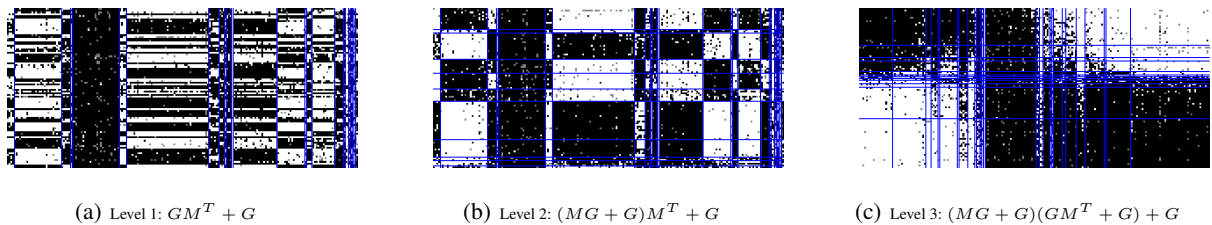

(a) Level 1: $GM^T + G$      (b) Level 2: $(MG + G)M^T + G$     (c) Level 3: $(MG + G)(GM^T + G) + G$

Figure 4: Visualization of the representations learned from the Senate voting data. Rows = Senators, columns = votes. 200 columns were selected at random from the original matrix. Black = yes, white = no, gray = absence. Blue = cluster boundaries. Rows and columns are sorted first by cluster (if applicable), then by the highest variance dimension of the low-rank representation (if applicable). Clusters are sorted by the same dimension as well. The models in the sequence increasingly reflect the polarization of the Senate.

What is the significance of the grammar being context-free? While it imposes no restriction on the models themselves, it has the effect that the grammar "overgenerates" model structures. Our grammar licenses some nonsensical models: for instance, $G(MG + G) + G$, which attempts to cluster dimensions of a latent space which is defined only up to affine transformation. Reassuringly, we have never observed such models being selected by our search procedure — a useful sanity check on the output of the algorithm. The only drawback is that the system wastes some time evaluating meaningless models. Just as context-free grammars for English can be augmented with attributes to enforce contextual restrictions such as agreement, our grammar could be similarly extended to rule out unidentifiable models. Such extensions may become important if our approach is applied to a much larger space of models.

Our context-free grammar formalism unifies a wide variety of matrix decomposition models in terms of compositional application of a few production rules. We exploited this compositional structure to efficiently and generically sample from and evaluate a wide variety of latent variable models, both continuous and discrete, flat and hierarchical. Greedy search over our grammar allows us to select a model structure from raw data by evaluating only a small fraction of all models. This search procedure was effective at recovering the correct structure for synthetic data and sensible structures for real-world data. More generally, we believe this paper is a proof-of-concept for the practicality of selecting complex model structures in a compositional manner. Since many model spaces other than matrix factorizations are compositional in nature, we hope to spur additional research on automatically searching large, compositional spaces of models.

# Acknowledgments

This work was partly funded by the ARO grant W911NF-08-1-0242 and by an NDSEG fellowship to RBG.